# Semantic Understanding of Professional Soccer Commentaries

**Hannaneh Hajishirzi, Mohammad Rastegari, Ali Farhadi, and Jessica K. Hodgins**
{hannaneh.hajishirzi, mrastega, afarhadi, jkh}@cs.cmu.edu
Disney Research Pittsburgh and Carnegie Mellon University

## Abstract

This paper presents a novel approach to the problem of semantic parsing via learning the correspondences between complex sentences and rich sets of events. Our main intuition is that correct correspondences tend to occur more frequently. Our model benefits from a discriminative notion of similarity to learn the correspondence between sentence and an event and a ranking machinery that scores the popularity of each correspondence. Our method can discover a group of events (called *macro-events*) that best describes a sentence. We evaluate our method on our novel dataset of professional soccer commentaries. The empirical results show that our method significantly outperforms the state-of-the-art.

## 1 Introduction

This paper addresses the problem of understanding professional commentaries of soccer games. Computational understanding of such domains has crucial impact in automatic generation of commentaries and also in game analysis and strategic planning. To this end, one needs to infer the semantics of natural language text; this is an extremely challenging problem. Understanding professional soccer commentaries further introduces interesting and challenging issues. For example, commentators do not typically talk about all the events of the game, selecting what is important. Also, they use a variety of phrases to report similar events. For example, a simple event of "A passes to B" can be commentated in several different ways: "A feeds B", "A and B in a nice combination", "A, what a beautiful way to B". Further, in some cases, commentators create a group of events and only mention a *macro-event*. For example, instead of saying "A passes to B, B passes to C, and C passes to D", the commentators report this whole sequence of events as "Team X is coming forward" or "nice attack by X". Also, professional commentators report several statistics and related information about the league, players, stadium, and weather during less interesting segments of a play.

A general solution to understanding such a complex phenomenon requires infering about game-related events, reasoning in terms of very complex paraphrases, and also forming high-level understandings of game events. Most recent work in semantic parsing of natural language translates individual sentences into the underlying meaning representations. Meaning representations are usually logical forms represented with events or relations among entities. The problem of semantic parsing can be formulated as learning to map between sentences and meaning representation in a supervised fashion [Zettlemoyer and Collins, 2005]. One can decrease the amount of supervision in specific controlled domains, such as RoboCup soccer [Chen *et al.*, 2010; Hajishirzi *et al.*, 2011] and Windows help instructions [Branavan *et al.*, 2009]. Recently, [Liang *et al.*, 2009] introduce a general semantic parsing technique that is not restricted to a specific domain, but is not scalable to large datasets due to the complexity of the model. In this paper, we introduce an algorithm that does not require domain-specific knowledge and is scalable to larger datasets.

We formulate the problem of understanding soccer commentaries as learning to align sentences in commentaries to a list of events in the corresponding soccer game. Our approach does not need expensive supervision in terms of correspondences between sentences and events. Similar to previous work [Liang *et al.*, 2009; Chen *et al.*, 2010], we use loose temporal alignments between sentences in commentaries and events of games. We pair sentences with several events that occur in the rough temporal vicinities of the sentences. Each pair consists of a sentence and a corresponding event. We then try to distinguish between correct and incorrect pairs. We rank pairs based on how consistently they appear in other places. We use a discriminative notion of similarity to reason about repetitions of pairs of sentences and events. The core intuition is that, under appropriate similarity metrics, correct pairs of sentences and events appear more often across several games. Our model is capable of forming macro-events and matching a group of events to a sentence. Experimental evaluations show that our method significantly outperforms the state of the art in our dataset and a benchmark of RoboCup soccer dataset. Our qualitative results demonstrate how our method can form macro-events and reason about complex paraphrases. We also introduce a dataset of eight English Primier League games with their commentaries and the accurate log of the game events. For evaluation purposes, we manually aligned sentences with events. Our dataset is publicly available.

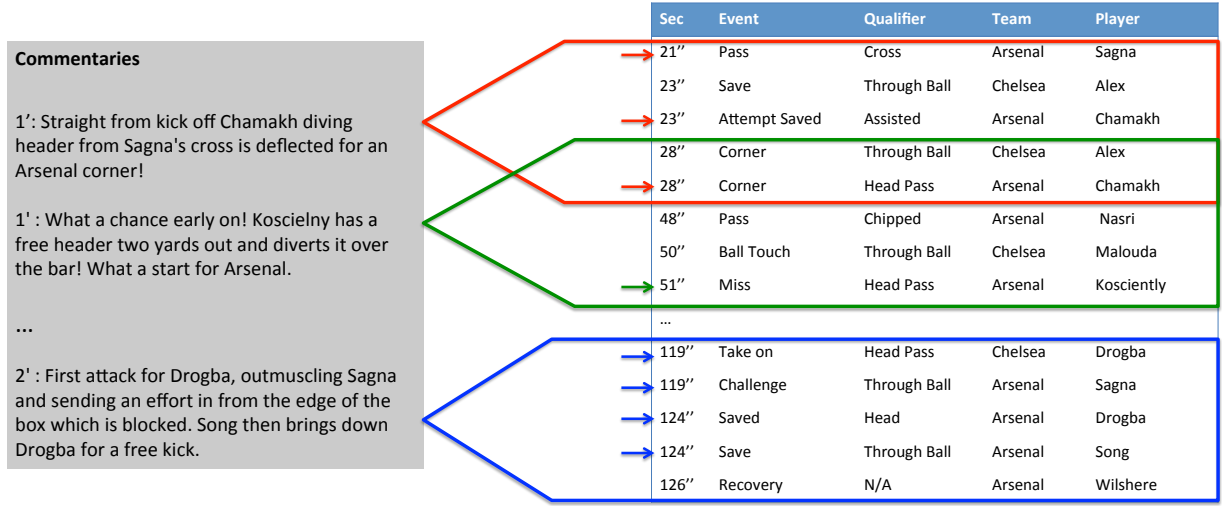

Figure 1: Examples of sentences in the commentaries with corresponding buckets of events. The correct correspondences in each bucket are marked with arrows.

### 1.1 Related Works

There are several approaches that learn to map natural language texts to meaning representations. Some researchers use full supervision in general and controlled domains [Zettlemoyer and Collins, 2005; Ge and Mooney, 2006; Snyder and Barzilay, 2007] . More recent work reduces the amount of required supervision in mapping sentences to meaning representations in controlled domains. [Kate and Mooney, 2007] introduce an Expectation Maximization (EM) approach with weak supervision in the GeoQuery and Child corpora. [Branavan *et al.*, 2009; Vogel and Jurafsky, 2010] use a reinforcement learning approach to map Windows help or navigational instructions to real events in the world. The supervision is weakly provided through interactions with a physical environment. [Chen *et al.*, 2010; Hajishirzi *et al.*, 2011] introduce EM-like approaches to map RoboCup soccer commentaries to real events of the game. They use weak supervision through the temporal alignments between sentences and events or the domain knowledge about soccer events. [Poon and Domingos, 2009] introduce an unsupervised method that finds is-a relationships among the entities in the domain by taking advantage of clustering structures of sentences and arguments. Most previous work take advantage of the special properties of the underlying domains and cannot be applied to the commentaries of real soccer games due to the complexity of the domain.

The closest work to ours are [Liang *et al.*, 2009; Bordes *et al.*, 2010] that introduce an interesting generative approach and a ranking system to map texts to meaning representations, respectively. The generative approach of [Liang *et al.*, 2009] is general, but it is not scalable as we demonstrate in our experiments. The system of [Bordes *et al.*, 2010] uses different models and objectives for ranking functions. Also, these two methods, as well as most previous work, has a one-to-one restriction in mapping between segments and events. This is not a well-suited assumption for real-world domains like professional soccer commentaries. One exception is the work of [Branavan *et al.*, 2010] that finds high-level comments which is specific to Windows instructions.

## 2 Problem Definition

The input to our problem is a representation of temporal evolution of the world state (sequence of events) and a natural language text (sequence of sentences). We use a rough alignment between every sentence $S_i$ and a list of events $B(S_i)$; we call this list of events a *bucket* associated with the sentence $S_i$. We form a *pair* by combining a sentence $S_i$ with an event in the corresponding bucket $B(S_i)$. The sentence $S_i$ may describe zero, one, or more events in the bucket $B(S_i)$. Our objective is to find events in $B(S_i)$ that correspond to the sentence $S_i$. Figure 1 shows three example sentences and a portion of corresponding buckets in the dataset. Due to very loose alignments between sentences and events one has to consider a rather big bucket size.

Every event $e(\vec{x})$ is represented by an event type $e$ and a list of arguments $\vec{x}$. The arguments of an event can take two different types: string and categorical. String arguments correspond to lexical properties, such as player names, that can be easily found through a lexical analysis. Often only a part of the argument appears in the text. For instance, "Steven Gerrard" might be referred as "Gerrard" or "Steven". Categorical arguments correspond to discrete values in the domain such as zone(back, left, front, right). These arguments might be addressed in many different ways and cannot be identified with lexical search. In fact, identical words to the categorical values do not convey any information. For ease of notations, we call each event $e_j$, discarding all the arguments.

**Professional Soccer Commentaries Dataset (PSC)** This paper introduces a dataset of professional soccer commentaries aligned with real events that occur in each match. We also provide ground-truth alignments between sentences in the commentaries and events of the game for evaluation purposes only. Our model does not use any domain specific knowledge, but we focus on soccer domain in this paper.

**Commentaries:** Commentaries are English texts, generated by professional soccer commentators. We collect the professionally generated commentaries for matches in the 2010-2011 season of English Premier League from Espn.net.

**Soccer Events:** The events are all actions that occur around the ball and are labeled by human annotators. We use the F24 soccer data feed collected for the EPL by Opta [Opta, 2012]. The F24 data is a time-coded feed that lists all events within the game with a player, team, event type, minute, and second for each event. Each event has a type together with a series of arguments describing it. The data include different types of events, such as goals, shots, passes (with start/end point), tackles, clearances, cards, free kicks, corners, offsides, substitutions and stoppages. This data is currently used for real-time online visualizations of events, as well as post-analysis for prominent television shows and newspaper articles.

**Ground Truth Alignments:** For evaluation purposes only, we generate the ground truth labels which denote the correct mapping between sentences in the commentary and events in the F24 Opta feed. Every sentence can be matched with zero, one, or more events in the bucket.

The commentaries, list of events with their time stamp, and the features are publicly available at: http://vision.ri.cmu.edu/data-sets/psc/psc.html.

This domain is much more complex than the RoboCup soccer, weather, or Windows instruction datasets used by previous work. The sentences are more complex. The commentator uses different phrases to refer to an identical event type. Most of the arguments in the events are categorical, which cannot be aligned using lexical analysis. For some buckets, there are no events that correspond to the sentence because the sentence is either a general statement about the game or is about game statistics, weather, etc. For many buckets, multiple events correspond to the sentence. Figure 3 shows examples of sentences with 2, 3, or 4 corresponding events that our approach finds correctly.

## 3 Approach

The problem of semantic understanding of commentaries can be formulated as establishing alignments between a sentence and a set of events. This, in fact, reduces to figuring out which events in the bucket $B(S_i)$ correspond to the sentence $S_i$. Every sentence can be aligned with multiple events in the corresponding bucket. Providing supervision at the level of sentence-event alignments is tedious and expensive. Instead, we use the weak supervision in the rough alignments between sentences and events.

To set up notations, assume that the input to our system includes $s$ sentences represented by $S_i, i \in \{1 \ldots s\}$, and each sentence $S_i$ corresponds to a bucket of events $B(S_i)$. Without loss of generality, we can assume that each bucket has $n$ events: $B(S_i) = \{e_1^i, e_2^i, ..., e_n^i\}$. Each sentence $S_i$ can be aligned with a group of events, meaning that each sentence $S_i$ can be matched to a member of the power set of $B(S_i)$, called $\mathcal{P}(B(S_i))$. The cardinality of this set is $2^{|B(S_i)|}$. The problem of finding a group of events that best aligns to the sentence $S_i$ can be formalized as:

$$\arg\max_{\mathcal{E}_i \in \mathcal{P}(B(S_i))} \rho(S_i, \mathcal{E}_i)$$
$$\mid \mathcal{E}_i \mid \leq k \quad (1)$$

where $\rho$ is a ranking function that scores *pairs* of sentences and events based on the quality of the correspondence and $\mathcal{E}$ is a group of events (called macro-event) of a cardinality less than or equal to $k$. This is obviously a search in an exponential space. Later, we show that Equation 1 is a form of budgeted submodular maximizations. We also show how we can search this space more efficiently.

Before going into the details of our search method, we need to specify a ranking function, $\rho$. The main role of $\rho$ is to score the quality of each pair; each pair includes a sentence and an event in the corresponding bucket. The core idea is that under an "appropriate" notion of similarity, a good pair tends to appear more consistently across the data. For that, one needs to somehow count the number of appearances of a pair. Simple counting does not work because exact sentences and events may not appear more than once. However, the underlying pattern of correspondence between a sentence and an event may appear rather frequently. Therefore, we need a way of capturing the patterns of correspondences for pairs and then score them based on the popularity of the patterns in the dataset.

## 3.1 Learning Pair Models

We adopt a discriminative approach that learns the characteristic correspondence patterns that distinguishes a pair from all other pairs. Recently in computer vision, Exemplar Support Vector Machines (ESVM) has shown great success in learning what is unique about an image that can distinguish it from all other images [Malisiewicz *et al.*, 2011; Shrivastava *et al.*, 2011]. The main idea is very simple, yet surprisingly effective. To learn what is unique about each example, one can fit an SVM with only one positive instance and large number of negative instances. The main intuition is that an example can be defined as what it is not, rather than what it is like. Despite being susceptible to overfitting, the proposed hard negative mining method gets away from this issue.

This framework suits our problem setting very well because we learn for all pairs of sentences and events in each bucket and do not need training labels for which pairs are correct. We extend this approach to learn models of pairs (called *PairModel*). A *PairModel* demonstrates how to weigh features of a pair against each other in a discriminative manner. If a learned model for a pair produces a positive score when applied to another pair, then two pairs share analogous patterns of correspondences.

**Feature Vector:** The features for each pair $p_{ij} = (S_i, e_j^i)$ of a sentence $S_i$ and event $e_j^i$ are $\vec{\Phi}_{ij} = (\vec{\Phi}_{S_i}, \vec{\Phi}_{e_j^i}, \vec{\Phi}_{st_i})$. $\vec{\Phi}_{S_i}$ is a binary vector representing the sentence $S_i$, where each element in the vector shows the presence of a word in the vocabulary. The vocabulary consists of frequent words in the domain except the words that can occur in the string arguments. For instance, the vocabulary does not include the `player names`. $\vec{\Phi}_{e_j^i}$ is a binary vector representation of the event $e_j^i$ that includes the event type together with its arguments. Each element in the vector represents the presence of the corresponding event type or the argument value in $e_j^i$. $\vec{\Phi}_{st_i}$ is a binary vector with one element for every string type in the event arguments. Every element in the vector $st_i$ denotes if the string argument is matched with a word in the sentence. For instance, the feature vector has an argument to demonstrate whether or not the `player name` (argument of the event $e_j^i$) has occurred in the sentence $S_i$.

**Pair Model:** For each pair $p_{ij} = (S_i, e_j^i)$, we fit a linear SVM to a set of pairs with $p_{ij} = (S_i, e_j^i)$ as a positive training example and a large number of pairs as negative examples. The negative examples are selected in a way that encourages weak similarities between patterns of correspondences compared to the positive example. To do that, we try to make sure that none of the examples in the negative set contains the identical sentence or event to the positive example. If the events (resp. sentences) are similar (in Euclidean distance) to the event (resp. sentence) of the positive example, we make sure that sentences (resp. events) are very different (more details in Algorithm 1 and Section 4.1). We balance the examples by weighting them accordingly. The algorithm is sketched in Algorithm 1.

The output of the *PairModel* $M_{ij}$ learned for the pair $p_{ij} = (S_i, e_j^i)$ is a weight vector $\vec{\Theta}_{ij}$. The confidence of applying $M_{ij}$ over a new pair $p_{kl} = (S_k, e_l^k)$ is computed as $Conf(M_{ij}, p_{kl}) = \vec{\Theta}_{ij} \cdot \vec{\Phi}_{kl}$. This confidence compares the patterns of correspondence between $S_i$ and $e_j^i$ and patterns of correspondence between $S_k$ and $e_l^k$. The *PairModel* tries to weight the features that are most important to discriminate the corresponding pair from the rest. High confidence means that those important features are "on" in an example; therefore, it is following the same pattern. Figure 2 demonstrates three *PairModels*, the top-weight words for each pair, and the nearest sentence under the learned patterns of correspondences. Non-discriminative measures of similarity like Cosine or Euclidean are not desirable because they treats all the dimensions in the same way. For comparisons and experimental evaluations please see Section 4.

A *PairModel* $M_{ij}$ *likes* the pair $p_{kl}$ by $Conf(M_{ij}, p_{kl})$. We score the similarity between two pairs by looking at their mutual *likeness* meaning that they both share analogous patterns of correspondence. We can now start reasoning about the popularity of *PairModels* by aggregating all mutual likeness scores. To propagate the mutual likeness information we adopt a strategy similar to Google PageRank.

## 3.2 Ranking Pairs

The input to this module are all of the pairs and their learned models, and the output is the popularity scores of the pairs relative to each other. A pair $p_{ij} = (S_i, e_j^i)$ is likely to be a correct match if the pattern of correspondence extracted by $M_{ij}$ occurs frequently relative to other pairs; this means that $p_{ij}$ is popular . A pair is frequent if many popular pairs like that pair with high confidence. This resembles similar problems in ranking Web pages.

We adopt an approach similar to the PageRank algorithm [Brin and Page, 1998] utilized by Google to order the importance of webpages. PageRank examines the graph of webpages (called Webgraph) and assigns a high score to a web page if many important pages link to the page.

Here, we are interested in computing relative scores of the pairs. We build a graph of pairs by assigning a node to each pair. Unlike the Webgraph, the edges are undirected and weighted. We assign a weighted edge between a pair $p_{ij}$ and a pair $p_{kl}$, if they mutually like each other (i.e., $Conf(M_{ij}, p_{kl}) > 0$ and $Conf(M_{kl}, p_{ij}) > 0$). We call this edge the *popularity* link. The graph also has self loops; i.e., a node can also be connected to itself if $Conf(M_{ij}, p_{ij}) > 0$. This self loop encodes how com-petent each *PairModel* is. The weight of an edge between pairs denotes the degree of confidence that each pair likes the other. Calibrating pairmodels against each other is an issue. For that, we use the rank of each pair among all the other pairs to model the degree of confidence. More formally, the weight of the edge between $p_{ij}$ and $p_{kl}$ is $1/(rank(M_{kl}, p_{ij}).rank(M_{ij}, p_{kl}))$ where $rank(M_{kl}, p_{ij})$ shows the order of $p_{ij}$ among all the pairs $p$ with $Conf(M_{ij}, p) > 0$.

Our approach, *PairRank* (sketched in Algorithm 3), first builds the adjacency matrix of the graph using the edges and their weights. The ranking function $\rho(p_{ij})$ iteratively computes the popularity score of a pair according to Equation 2. At every iteration, $\rho(p_{ij})$ is the expected sum (with probability $d$) of the score of the adjacent pairs (computed at the previous iteration) and the self confidence value:

$$\rho(p_{ij}) = (1-d)Conf(M_{ij}, p_{ij}) + d \sum_{p_{kl} \in T(p_{ij})} \frac{\rho(p_{kl})}{edge(p_{ij}, p_{kl})} \quad (2)$$

where $edge(p_{ij}, p_{kl}) = rank(M_{ij}, p_{kl}).rank(M_{kl}, p_{ij})$, $T(p_{ij})$ is the set of adjacent nodes to $p_{ij}$, $d$ is a damping factor, and $\rho(p_{ij})$ is initialized by random values.

At iteration 1, only popularity links with length 1 are considered; $\rho(p_{ij})$ only adds up the scores of the pairs that are directly linked to $p_{ij}$. In next iterations, longer popularity paths are considered; the effect of indirectly linked pairs to $p_{ij}$ is included in the scores of neighboring pairs. We are not interested in adding the effect of pairs with high distances from $p_{ij}$. We control the expected length of the popularity paths with a damping factor $d$.

At the end of each iteration, we divide $\rho(p_{ij})$ by the frequency of the type of the event $e_j^i$ to discount the biases in the dataset. For instance, there are 8,597 events with type `pass` in the dataset. However, only 451 (5.2%) of the `pass` events occur in the ground-truth events. The last step of each iteration is to normalize the scores of all the pairs.

### 3.3 Searching for Good Correspondences

Now that we learn *PairModels* and rank the pairs based on their popularity, we return to our main goal of aligning sentences with events. If we knew that each sentence only corresponds to one event we could report the arg max of the outputs of the *PairRank* scores. However, most of the times sentences correspond to groups of events merged together, called macro-events.

Dealing with macro events requires learning *PairModels* and performing the *PairRank* on pairs with macro-events. To pair a sentence $S$ with a macro-event $\mathcal{E}$ we need to define how to merge events to form macro-events. Assume that we want to merge $(S, e_1)$ and $(S, e_2)$ to form the pair $(S, \mathcal{E})$. A *PairModel* for $(S, \mathcal{E})$ should learn the correspondences between the sentence $S$ and all the events in $\mathcal{E}$. To learn the *PairModel* for $(S, \mathcal{E})$, we use $(S, e_1)$ and $(S, e_2)$ as positive examples and generate negative examples as before (Algorithm 1). This results in a *PairModel* $M$. The confidence of the *PairModel* $M$ over a pair $P_k = (S_k, \acute{\mathcal{E}})$ is the maximum confidence of the $M$ on pairs of the sentence with every event in the macro event $\acute{\mathcal{E}}$ (Algorithm 2). $Conf(M, P_k) = \max_{e_j^k} Conf(M, p_{kj} = (s_k, e_j^k))$.

To score the popularity of every pair with macro-event, we run the *PairRank* method by adding one node per each pair with macro-event. The edge weights to the new node are computed as the maximum score of all the events in the macro-event (Algorithm 2).

As mentioned before, the search space for each sentence is the exponential space of all possible ways of merging events in each bucket (Equation 1). However, the good news is that our main objective function in Equation 1 is submodular. Because by definition, the ranking function works as a maximization of a set function over the examples that form the macro-event. This results in $\rho(S_i, \mathcal{E}_j) + \rho(S_i, e_k) \geq \rho(S_i, (\mathcal{E}_j \oplus e_k)) + \rho(S_i, (\mathcal{E}_j \ominus e_k))$. We denote the operation of merging two events by $\oplus$ and the inverse operation by $\ominus$. These operators resemble the union and intersections over sets. Now we can adopt a greedy approximation of Equation 1 with reasonable error bounds [Goundan and Schulz, 2009; Dey *et al.*, 2012; Krause *et al.*, 2008]. The core intuition is that instead of searching the exponential space of all possible ways of merging events, we start with the best scoring event and merge events that maximizes the marginal benefits of merging them. We keep merging until we observe no benefit of doing so. More formally, our recursive greedy solution is:

$$\mathcal{E}^l = \mathcal{E}^{l-1} \oplus \mathcal{A}^*$$
$$\mathcal{A}^* = \arg \max_{A \in B(S) \setminus D} \rho(S, \mathcal{E}^l \oplus A) - \rho(S, \mathcal{E}^l) \quad (3)$$

where $\mathcal{E}^0 = \emptyset$, $\mathcal{E}^l$ is the macro-event, of cardinality at most $l$, that we want to grow using the merging procedure $\oplus$, and $D$ is the set of all elements in $\mathcal{E}^l$. We stop this procedure after $k$ steps or when merging events does not help (Algorithm 4).

To elaborate more, assume that for the sentence $S$ we are given the bucket of events $B(S) = \{e_1, e_2, e_3\}$. We start with the best scoring event in $B(S)$, let's say $e_1$. We then look for the event that maximizes the marginal benefit (Equation 3), let's say $e_2$. We now check to see if there is any gain in forming macro-events. If there is no gain we stop and report $e_1$ as the answer. Otherwise, we form the macro-event $e_{12} = e_1 \oplus e_2$ by merging the two events together, ($\mathcal{E}^2 = e_{12}$). We now can go to the next layer and search for the next event that maximizes the marginal benefit, let's say $e_3$. If adding $e_3$ helps, we form a new macro-event, $\mathcal{E}^3 = e_{123} = e_{12} \oplus e_3$. This procedure may result in aligning sentences with macro-events of cardinality up to $k$. In our experiments we set the $k = 4$.

**Algorithm 1.** *PairModel*$(S, \mathcal{E})$
- Input: pair $(S, \mathcal{E})$
1. $\vec{\Phi} \leftarrow$ feature vector for $(S, e_j) \; \forall e_j$ in $\mathcal{E}$
2. $Pos \leftarrow \cup_j (S, e_j) \; \forall e_j$ in $\mathcal{E}$
   // Generate Negative Examples
3. $\forall$ pairs: sort $D_e \leftarrow Dist(e, e_k)$, sort $D_S \leftarrow Dist(S, S_k)$
4. $Neg_1 \leftarrow (S_k, e_l) : S_k \in D_{S1:N}, e_l \in D_{e_{end-N:end}}$
5. $Neg_2 \leftarrow (S_k, e_l) : S_k \in D_{Send-N:end}, e_l \in D_{e_{1:N}}$
6. $Neg \leftarrow Neg_1 \cup Neg_2$
7. return $SVM(Pos, Neg)$ with the weight vector $\vec{\Theta}$

**Algorithm 2.** *ComputeConf*$(p_{ij}, p_{kl})$
- Input: pairs $p_{ij} = (S_i, \mathcal{E}_j^i)$ and $p_{kl} = (S_k, \mathcal{E}_l^k)$ with feature vector $\Phi_{kl}$
1. $M_{ij} \leftarrow PairModel(p_{ij})$ (Alg. 1)
2. $\vec{\Theta}_{ij} \leftarrow$ weight vector of $M_{ij}$
3. $Conf(M_{ij}, p_{kl}) \leftarrow \max_{e_l \in \mathcal{E}_l^k} \vec{\Theta}_{ij} . \vec{\Phi}_{kl}$

**Algorithm 3.** *PairRank*$(pairs)$
- Output: ranks of all the pairs
//Build adjacency matrix
1. for pairs $p_{ij}, p_{kl}$:
   (a) *ComputeConf*$(p_{ij}, p_{kl})$ (Alg. 2)
   (b) $\vec{P}_i \leftarrow$ scores of applying $M_{ij}$ on all pairs
   (c) $\vec{P}_k \leftarrow$ scores of applying $M_{kl}$ on all pairs
   (d) $edge(p_{ij}, p_{kl}) \leftarrow (rank(p_{kl})$ in $\vec{P}_i)$ and $(rank(p_{ij})$ in $\vec{P}_k)$
//Iteration $t$
2. while $|\vec{\rho}_t - \vec{\rho}_{t-1}| \geq \epsilon$
   (a) for every pair $p_{ij} = (S_i, e_j^i)$
       i. $\rho_t(p_{ij}) \leftarrow$ Equation 2
       ii. $\rho_t(p_{ij}) \leftarrow \frac{\rho_t(p_{ij})}{\text{frequency}(e_j^i.type)}$
       iii. $\rho_t(p_{ij}) \leftarrow \frac{\rho_t(p_{ij})}{sum(\vec{\rho})}$
3. return $\vec{\rho}_t$

**Algorithm 4.** *ReturnMacroEvent*$(pairs, k)$
- Output: best macro event for each bucket
1. Iteration $l = 0$:
   (a) $\mathcal{E} = \emptyset$
   (b) $\vec{\rho} \leftarrow PairRank(pairs)$
   (c) $\mathcal{A}^* \leftarrow$ highest ranked pair in each bucket according to $\vec{\rho}$
2. Iteration $l$:
   (a) $\mathcal{E} \leftarrow \mathcal{E} \oplus \mathcal{A}^*$
   (b) $\vec{\rho}_1 \leftarrow PairRank(pairs \cup \mathcal{A}^*)$
   (c) for $e_i$ in every bucket:
       i. $\vec{\rho}_2 \leftarrow PairRank(pairs \cup \mathcal{E} \oplus e_i)$
       ii. $\Delta_{e_i} \leftarrow \rho_2(S, \mathcal{E} \oplus e_i) - \rho_1(S, \mathcal{E})$
   (d) $\mathcal{A}^* \leftarrow \arg\max_{e_i} \vec{\Delta}$
3. repeat until $l \leq k$
4. return $\mathcal{E}$ for each bucket

## 4 Experiments

We evaluate our method on how accurately it aligns sentences in professional soccer commentaries with events in the actual games. We use our professional soccer dataset as the main testbed and compare our method with state-of-the-art methods and several different baselines. For comparison, we also test our model on a benchmark dataset of RoboCup soccer [Chen and Mooney, 2008].

### 4.1 Professional Soccer Commentaries

Our dataset consists of time-stamped commentaries and event logs of 8 games in 2010-2011 season of English Premier League. For evaluation purposes, we label ground-truth annotations for the correspondences between sentences and events throughout the dataset. There are 935 sentences, 14,845 events, 2,147 words, and 306 players in total. Each sentence on average has 16.62 words. For each sentence in the commentaries, we assign a bucket by selecting events that occur in an interval of 150 seconds around the time the sentence has been generated. We then pair each sentence with all of the events in the corresponding bucket. This results in 38,332 pairs. On average, there are 42 pairs in each bucket. Of course, not all of these pairs are correct correspondences. There are in total 1,404 pairs labeled as correct matches in the ground-truth labels. On average there are 2 correct pairs in each bucket.

Each event is represented with an event type followed by a list of its arguments. There are 55 event types and several arguments defined in the event logs. Examples of the arguments are the `time` that the event occurred, the `player name` that is the agent of the event, the `team name`, the `outcome` of the event, and the `body part`. Most of the arguments are categorical. The only string field is the `player name` that can provide useful information for finding initial guesses for correct correspondences. However, identical player names occur in multiple events in each bucket.

We apply our method (Algorithm 4) on this dataset. We start by pairing sentences with events in the corresponding buckets. This step is followed by learning *PairModels* for every pair $p_{ij} = (S_i, e_j^i)$ that has at least one matching player name between the sentence $S_i$ and the event $e_j^i$. The features for the pair $p_{ij}$ are $\Phi_{ij} = (\Phi_{S_i}, \Phi_{e_j^i}, \Phi_{st})$ where $\Phi_{S_i}$ and $\Phi_{e_j^i}$ are binary vectors representing the sentence and the event, and $\Phi_{st}$ is an integer value corresponding to the number of players matched between the sentence and 3 consecutive events. At the first step, each *PairModel* $M_{ij}$ is trained using one positive example $(S_i, e_j^i)$ and 100 negative examples generated automatically as follows. We sort all sentences according to their Euclidean distance to the sentence $S_i$ and all events based on their Euclidean distance to the event $e_j^i$. We generate negative examples from the sentences and events that are not identical to $S_i$ and $e_j^i$. Half of the negative pairs are generated from the sentences with lowest distance to $S_i$ and events with highest distance to $e_j^i$. The other half are generated by pairing sentences with highest distance to $S_i$ and events with lowest distance to $e_j^i$. We make sure that no pairs with matching player names between sentences and events exist in the negative set. This way we try to make sure that negative examples do not contain the same patterns of correspondences as the positive examples.

| Sentence | Chelsea looking for a penalty as Malouda's header hits Koscielny, not a chance as it hit him in the stomach. | First attack for Drogba, outmuscling Sagna and sending an effort in from the edge of the box which is blocked, and Song then brings down Drogba for a free kick | Two poor efforts for the price of one from the free kick as Nasri's shot hits the wall before Vermaelen fires wide. |
|---|---|---|---|
| Event | **E:**Foul **Q:**Head Pass **T:**Chelsea **P:**Malouda | **E:**Challenge **Q:** Through Ball **T:**Arsenal **P:**Sagna | **E:**Miss **Q:** Pass **T:**Arsenal **P:**Sagna |
| Top Weights | 'penalty', 'looking', 'hits', 'Foul', 'header', 'chance', 'chelsea', 'city', 'clear','half' | 'box', 'first','edge','block','attack','effort', 'kick', 'free','wide', break | 'shot', 'wide', 'two', 'hits', 'poor','one' 'free', 'kick', 'Miss', 'chelsea' |
| Top Sentences | Alex Song is too strong for Malouda, who goes down looking for a foul. Nothing given by the referee. | Foul by Arshavin as he blocks Essien's attack. He's lucky to escape a card there. | A choppy start from both sides with possession swapping hands regularly. Wilshere gets stuck into Jovanovic and gives away a free kick. |

Figure 2: Qualitative analysis of *PairModels*: Each column corresponds to a *PairModel* trained with a sentence on the first row and the event in the second row as the positive example. The learned *PairModel* assigns high values to the features corresponding to the words in the third row. The closest sentence under the learned pattern of correspondences between the sentence and the event in the pair is shown in the fourth row.

We then fit linear SVMs [Fan *et al.*, 2008] to the positive and negative examples for each pair. We use LibLinear with $c = 100$. We weight positive instances to avoid the affects of unbalanced data. We then apply the *PairRank* with the damping factor $d = 0.5$ to rank the pairs based on their consistency. To create macro-events we run Algorithm 4 with $k = 4$.

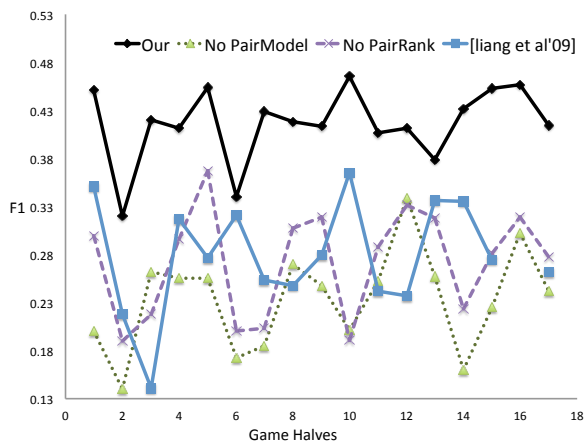

Figure 4: Our method outperforms the state-of-the-art on all the games.

| Method | $F_1$ | AUC | Precision | Recall |
|---|---|---|---|---|
| No PairModel | 23.3 | 36.4 | 37.3 | 17.1 |
| No PairRank | 27.3 | 37.4 | **39.4** | 21.1 |
| MIL | 11.0 | 36.8 | 37.3 | 7.0 |
| [Liang *et al.*, 2009] | 27.6 | N/A | 27.2 | 28.4 |
| Our approach | **41.4** | **46.8** | 33.9 | **54.0** |

Table 1: Average performance of different approaches over all games in our dataset.

### 4.1.1 Comparisons

We compare the performance of our method with the stat-of-the-art method of [Liang *et al.*, 2009], a Multiple Instance Learning (MIL) method of [Andrews *et al.*, 2002], and two baselines. As the main testbed, we use our professional soccer dataset where we have 16 half games (8 games). Figure 4 plots the comparisons mentioned above on all 16 half games. We use $F_1$ as the measure of performance per half game. Table 1 contains the results of all the aforementioned methods using $F_1$ and $AUC$ measures averaged over all half games.

**[Liang *et al.*, 2009]** This approach uses a generative model that learns the correspondence between sentences and events. We use their publicly available code and run their method for 5 iterations. To have a generous comparison, we report the best results achieved during 5 iterations (sometimes the best performance is achieved earlier than 5 iterations). Table 1 shows the average accuracy of this method over all the games. Our model outperforms this method by more than $14\%$ in $F_1$. Due to the complexity of the model, [Liang *et al.*, 2009] cannot take advantage of the full capacity of the domain. It runs out of memory on a machine with $8GB$ of memory when using 6 arguments. The reason is that this approach grows exponentially with the number of arguments whereas our approach grows linearly. To be compatible, we decrease the number of arguments of every event to 3 arguments `team`, `player-name`, `qualifier`, in all the comparisons. Even after decreasing the number of arguments, the method of [Liang *et al.*, 2009] still runs out of memory for one game that consist of long sentences (half-game 16). Due to memory limitations, this method is not applicable to all games at the same time; we perform all comparisons on half game basis.

**Multiple Instance Learning (MIL):** Finding correct correspondences given the rough alignments between sentences and events can be formulated as a multiple instance learning problem. Pairs of sentences with all the events in the corresponding bucket can be considered as bags. A positive bag includes at least one correct pair. Negative bags do not include any correct pairs. We utilize the same procedure that we used to generate negative pairs to produce neg-ative bags. We use the publicly available implementation of mi-SVM [Andrews *et al.*, 2002] for comparisons. Table 1 compares the performance of our method with that of MIL. Our method outperforms MIL significantly (by about 30% in $F_1$). We postulate that, the complex structure in the correspondences between sentences and events cannot be discovered by latent methods like mi-SVM.

| Sentences | Discovered Events | | | |
|---|---|---|---|---|
| 1: Chelsea looking for penalty as Malouda's header hits Koscielny, not a chance as it hit him in the stomach. | **E:** Pass | **Q:** Head Pass | **T:** Chelsea | **P:** F. Malouda |
| | **E:** Ball touch | **Q:** Through ball | **T:** Arsenal | **P:** L. Koscielny |
| | **E:** Foul | **Q:** through ball | **T:** Arsenal | **P:** L. Koscielny |
| | **E:** Foul | **Q:** Head Pass | **T:** Chelsea | **P:** F. Malouda |
| 2: GOAL, Drogba opens the scoring! Ramires finds Ashley Cole with a perfectly waited pass, Cole's low cross finds Drogba at the near post who back-heels it beyond a stranded Fabianski. 1-0 to the champions. | **E:** Pass | **Q:** Cross | **T:** Chelsea | **P:** A. Cole |
| | **E:** Goal | **Q:** Head Pass | **T:** Chelsea | **P:** D. Drogba |
| 3: Essien dispossesses Arshavin and earns Chelsea's fourth corner of the match | **E:** Intercept | **Q:** Cross | **T:** Chelsea | **P:** M. Essien |
| | **E:** Pass | **Q:** Pass | **T:** Chelsea | **P:** M. Essien |
| | **E:** Corner Awarded | **Q:** Pass | **T:** Chelsea | **P:** M. Essien |
| 4: Cole is sent off for a lunge on Koscielny, it was poor, it was late but I'm not entirely sure that should have been red. | **E:** Foul | **Q:** Long Ball | **T:** Liverpool | **P:** J. Cole |
| | **E:** Foul | **Q:** Through Ball | **T:** Arsenal | **P:** L. Koscielny |
| | **E:** Card | **Q:** None | **T:** Liverpool | **P:** J. Cole |
| 5: First attack for Drogba, outmuscling Sagna and sending an effort in from the edge of the box which is blocked, and Song then brings down Drogba for a free kick. | **E:** Take On | **Q:** Head Pass | **T:** Chelsea | **P:** D. Drogba |
| | **E:** Challenge | **Q:** Through Ball | **T:** Arsenal | **P:** B. Sagna |
| | **E:** Save | **Q:** Through Ball | **T:** Arsenal | **P:** A. Song |
| | **E:** Foul | **Q:** Through Ball | **T:** Arsenal | **P:** A. Song |
| 6: Two poor efforts for the price of one from the free kick as Nasri's shot hits the wall before Vermaelen fires wide. | **E:** Attempt Saved | **Q:** Head Pass | **T:** Arsenal | **P:** S. Nasri |
| | **E:** Miss | **Q:** Head Pass | **T:** Arsenal | **P:** T. Vermaelen |

Figure 3: Qualitative examples of macro-events discovered by our method in correspondence to sentences. For instance, in sentence 4 the commentator is implicitly describing a foul occurred in the game which led to a card but there is no explicit mention of 'Foul' nor 'Card' in the sentence. However, our method can discover both events correctly.

**No PairModel Baseline:** To analyze the importance of discriminative notion of correspondences, we also compare our method with a baseline that uses a non-discriminative notion of correspondences between sentences and events. In our model, we replace the *PairModel* with a non-discriminative similarity metric (such as Cosine or Euclidean). In our experiments, we did not find any considerable difference between the non-discriminative distances. Table 1 shows performance numbers using Euclidean distance. Replacing the *PairModel* with non-discriminative distances decreases the performance by 16%.

**No PairRank Baseline:** We also replace the PairRank component in our model with a voting scheme. In this baseline, we compute the confidence scores of applying all *PairModels* on all pairs. We then compute the score of each pair by counting the number of non-negative confidence values. Replacing the *PairRank* model with a voting scheme decreases the performance by 13%.

We also analyze the importance of incorporating macro-events. To this purpose, we enforce our model to produce macro-events at multiple different lengths. Figure 5 plots the $F_1$ measures against the maximum cardinality of macro events. Forming macro-events up to the cardinality 4 dramatically boosts the performance. Our experiments do not show any boost by adding longer macro-events.

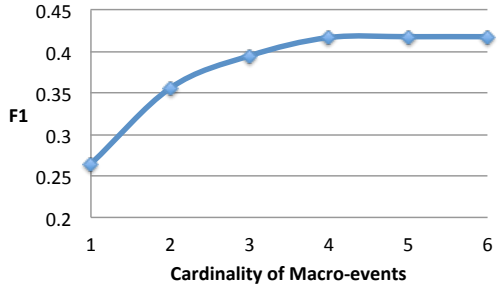

Figure 5: Average $F_1$ over all games by increasing the number of iterations.

| Approach | $F_1$ |
|---|---|
| [Chen and Mooney, 2008] | 67.0 |
| [Chen *et al.*, 2010] | 73.5 |
| [Liang *et al.*, 2009] | 75.7 |
| [Hajishirzi *et al.*, 2011] | 77.9 |
| Our approach | **81.6** |

Table 2: The average $F_1$ scores of 4 fold cross validation in the Roboup dataset.

#### 4.1.2 Qualitative Examples

Figure 3 shows examples of macro-events discovered by our method in correspondence to the sentences in the first column. The correspondences are not obvious and show deep understanding of sentences in commentaries. To further analyze the intermediate result of our model, we looked at the top scoring dimensions in the features for specific events. For example for event Foul, the highest weights correspond to 'Foul','free','kick','card', 'dangerous', and 'late'. For the event Save, the highest dimensions corresponds to 'Save','goal', 'shot','block','ball'.

### 4.2 RoboCup Soccer Commentary

We also compare our approach with the state-of-the-art methods in the RoboCup soccer dataset [Chen and Mooney, 2008]. The data is based on commentaries of four championship games of the RoboCup simulation league. Each game is associated with a sequence of human comments in English and the Robocup soccer events that happen in the original game tagged with time. Sentences are on average 5.7 words long. There are 17 events, including actions with the ball or other game information. The buckets are generated by mapping sentences and the events that occurred within 5 time steps of when the comments were recorded. There are in total 1,919 pairs and average of 2.4 events per bucket. It is assumed that every sentence is matched with at most one event. Following previous approach, we adopt the scheme of 4 fold cross validation and report the micro-averaged results for four games in terms of $F_1$. Table 2 demonstrates that our method outperforms the state-of-the-art. [Hajishirzi *et al.*, 2011] uses domain knowledge about soccer events instead of buckets information. [Chen *et al.*, 2010] can achieve the $F_1$ of 79.1% by initializing with the output generated by [Liang *et al.*, 2009]. We cannot directly compare [Bordes *et al.*, 2010] to other methods in the table because a) they use Bigram and Trigram features. This gives a strong boost for games like the game 3 where most sentences are 3 or 4 words long. b) they use post processing heuristics to find sentences with no matching events. With the same heuristics our model can go up to an F1 of 84.57 (cf. 83.0 of [Bordes *et al.*, 2010]).

## 5 Discussion and Future Work

In this paper, we present an approach that can form macro-events that best describe sentences given a bucket of events that correspond to each sentence. Our method takes advantage of a discriminative notion of correspondence coupled with a ranking technique to find popular pairs in our professional soccer dataset. To avoid exponential searches over all possible ways of merging events in a bucket we use a greedy approximation with reasonable bounds. We update the macro-events incrementally by combining events that provide maximum marginal benefits. Our experiments show significant improvement over the state-of-the-art methods. In fact, our method achieves an $F_1$ measure of 41.4% comparing to the state of the art performance of 27.6% in professional soccer commentaries. Also, our method outperforms state-of-the-art on RoboCup dataset.

Our method can assign a macro-event to a part of the text that cannot be further segmented. For instance, the first part of the sentence 1 in Figure 3 is mapped to both pass and corner. We argue that we need to first align sentences with macro-events and then start the segmentation with a relaxed one-to-one assumption. Discovering strategy-level macro events still remains open. For example our method cannot align attack or coming forward with series of passes. We believe that we can make progress by augmenting our method with more powerful Natural Language Processing tools that improves lexical analysis. Also, our method cannot address the cases where there is no event corresponding to a sentence. Our formulation forces at least one event per sentence. This is not desirable for many domains including soccer commentaries where there are several sentences that do not correspond to any ball-related event in the game. Our ultimate goal is to build an automatic commentary generation for professional soccer games. This paper is one step forward toward that big goal.

**Acknowledgments**

The authors would like to thank Opta for providing the data on event logs of the soccer games, Emma Brunskill for the useful discussions, and anonymous reviewers for their insightful comments on the paper. Ali Farhadi is supported by the ONR-MURI grant N000141010934.